\documentclass[final,5p,times,twocolumn]{elsarticle}

\usepackage{graphicx}
\usepackage{amssymb}
\usepackage{lineno}

\usepackage[colorlinks]{hyperref}

\usepackage{pdfpages}

\begin{document}

\begin{frontmatter}


\title{Analyzing Human-Human Interactions: A Survey}

\author{Alexandros Stergiou}
\ead{a.g.stergiou@uu.nl}
\address{Department of Information and Computing Sciences, Utrecht University,Buys Ballotgebouw, Princetonplein 5, Utrecht, 3584CC, Netherlands\fnref{label3}}

\author{Ronald Poppe}
\address{Department of Information and Computing Sciences, Utrecht University,Buys Ballotgebouw, Princetonplein 5, Utrecht, 3584CC, Netherlands\fnref{label3}}


\begin{abstract}
Many videos depict people, and it is their interactions that inform us of their activities, relation to one another and the cultural and social setting. With advances in human action recognition, researchers have begun to address the automated recognition of these human-human interactions from video. The main challenges stem from dealing with the considerable variation in recording setting, the appearance of the people depicted and the coordinated performance of their interaction. This survey provides a summary of these challenges and datasets to address these, followed by an in-depth discussion of relevant vision-based recognition and detection methods. We focus on recent, promising work based on deep learning and convolutional neural networks (CNNs). Finally, we outline directions to overcome the limitations of the current state-of-the-art to analyze and, eventually, understand social human actions.
\end{abstract}

\end{frontmatter}

%\linenumbers

\section{Introduction}
\label{Introduction}
Despite significant research progress in the automated analysis of humans and their activities \citep{cheng2015advances,herath2017going,koohzadi2017survey,poppe2010survey}, the recognition of human interactions from video remains a challenging topic. Integral part of the difficulty is that understanding interactions between people requires more than analyzing the actions of each person in isolation. Rather, it is the coordination, in both space and time, between people that reveals the true nature of their collective behavior. In addition, the context in terms of who is interacting why and where determines to a large extent how the interaction unfolds.

There is a long history of the manual and automatic description of human interactions, see \citep{birdwhistell1952introduction,poppe2017automatic,vinciarelli2009socialsignal} for overviews. Still, the relation between the observable form of the bodily interaction and the more subjective \textit{interpretation} thereof is relatively understudied. For example, putting a hand on someone's shoulder can be objectively identified, whereas more information is required to know that one person is comforting the other, or trying to get the other's attention. The scarcity of a more social, contextual perspective in the automated analysis of human-human interactions is also reflected in computer vision literature, where interactions are typically reduced to visually and temporally well-defined events. Despite this somewhat artificial view on human behavior, current advances pave the way for a more social perspective. In this paper, we survey the research in the recognition of human-human interactions in videos, with a focus on methods based on convolutional neural networks (CNNs). We then discuss promising directions to leverage the current state-of-the-art to a more social analysis.

\subsection{Scope and motivation}
\begin{figure}[!t]
\centering
\includegraphics[width=\columnwidth]{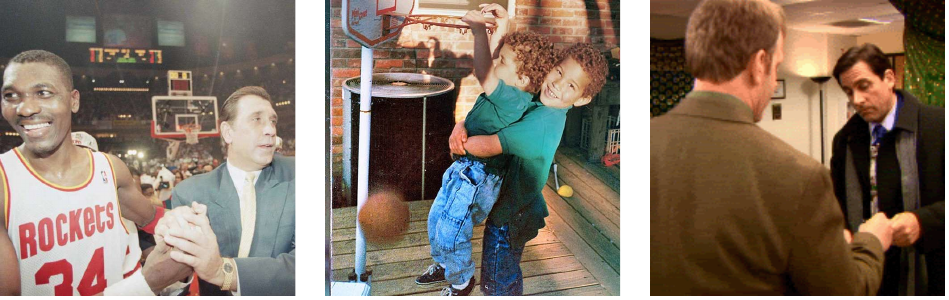}
\caption{Three interactions: handshake, hug or lift, and object passing. These examples show non-standard body poses (left), ambiguous class labeling (center) and the need for temporal information (right).}
\label{Figure1}
\end{figure}

% scope
In this survey, we focus on \textit{dyadic} interactions between two people. We consider joint actions of both people that can be characterized by the positions, movements and coordination of their bodies (see Figure~\ref{Figure1}). For example, we consider a handshake as an interaction that can be part of an \textit{activity} such as an agreement or a greeting. Interactions can be made up of several motions in sequence, such as extending the right arm, grasping the right hand of the other and moving the hands up-and-down. The duration of the interactions that we consider can be anywhere between half of a second and several seconds. There can be considerable variation in the performance of an interaction, most notably in the duration but also in the coordination. This variation can also lead to ambiguities in how they are perceived. For example, the hug interaction in Figure~\ref{Figure1}(center) could also be considered a lift interaction. The works discussed in this survey exclusively treat the interaction recognition task as deterministic, which does not fully reflect the more ambiguous nature in the perception of social behavior. We discuss alternative representations and methods in the \hyperref[Discussion]{Discussion} section.

% motivation
The automated recognition of bodily interactions from video mainly benefits content-based video retrieval \citep{liu2015content,sempena2011human}, security \citep{aran2013one} and surveillance \citep{cristani2011social,tian2012hierarchical,yi2014l0} and interactive human-computer interfaces \citep{rehg2013decoding,sheerman2011cultural}. The vast majority of the research has considered a functional perspective by labeling the visual aspect of videos. This leaves room for a more contextual interpretation of the joint behavior. Opportunities for a broader use of automated measures arise when computers can \textit{understand} the interactions in terms of communicative and affective intent. In this survey, we present the current basis and potential directions to take the important step from interaction recognition to understanding. We discuss the evolution of the current state-of-the-art in interaction recognition towards this \textit{social} perspective in the \hyperref[Discussion]{Discussion} section.

\subsection{Main challenges in the field} \label{Section1.2}
% section structure
We identify challenges when dealing with the visual and structural aspects of interaction videos. Additionally, we outline practical challenges in the development of methods of automated human-human action recognition.

\subsubsection{Variation in visual appearance}
% viewpoint
Interactions between people can be observed in many different environments, and under vastly different recording settings. Most notably, a change of viewpoint has a large effect on how the interaction is observed. Especially when people are interacting physically, it is likely that their body parts partially occlude each other. This presents challenges in the recognition of interactions from a single viewpoint, as characteristic movements or the poses of key body parts are not visible. Typically, we do not have access to other viewpoints to deal with potential ambiguities.

% appearance
Variation in clothing and lighting conditions further adds to the challenge of robustly observing the smaller movements. Especially in low-resolution videos, the level of detail might be insufficient to distinguish between subtly different interaction classes such as handshake and fist bump greetings.

\subsubsection{Intra-class variation in interaction performance}
% duration
The performance of an interaction in terms of body movements and coordination can differ significantly, see Figure~\ref{Figure1}(left). \citet{ronchi2015describing} has analyzed the variation for single images. Additionally, there is significant variation in the temporal execution of the movement. While such deviations can be used to differentiate between classes \citep{anderson2014toward}, the dissimilarity of performance within an interaction class is typically too large to derive general rules.

% sequential nature
Interactions, like individual actions, often present an intrinsic sequential nature of movements. For example, an extension of the hand of one person is normally followed with the extension of the other actor's hand. Results from works that aim at the prediction of future actions have immediate impact on the improvement of scene understanding (e.g., \cite{vondrick2016anticipating}). Other works build on the key idea that future actions can be predicted by classifying an action or interaction solely on its start \citep{ziaeefard2015time}. Such an approach might work well for goal-directed interactions \citep{cao2013recognize,ryoo2011human}, but is less successful when the variation in the performance increases (see Figure~\ref{Figure1}(right)). This is especially true when the interactions are more social and reactive in a communicative or affective way, such as jokingly stomping someone.

% tracking
Some works have addressed the estimation of a skeletal representation in order to circumvent having to learn interaction patterns directly from video \citep{cavazza2016kernelized,Pham2018,yub2015random,yun2012two}. Recent methods rely on CNN-based approaches (e.g., \citet{cao2017realtimepose,carreira2016human,guler2018densepose,insafutdinov2017arttrack,li2015maximum,Yang_2017_ICCV}) and allow to investigate both pose and movement of a person. Skeleton representations are informative for actions and interactions and present an attractive alternative or complement for image features. However, errors and inaccuracies in the pose estimation process might be propagated to the classification task. In addition, there is a need for quantitative units that capture the characteristic information of an interaction in terms of pose, movement and coordination in space and time.

\subsubsection{Challenges in data collection and labeling}
% practical challenges of the domain
The study of interactions is further complicated by a relative lack of large datasets. In Section~\ref{Section2}, we discuss the most popular resources, but most of them focus on a relatively limited domain (e.g. sports or surveillance). In addition, there is no common labeling of the interaction classes. For example, a handshake might be a category of its own, or might be part of a greeting class. This lack of standardization hinders cross-dataset studies and consequently limits the generalization of methods developed in one particular scenario to address another. While human-human interactions are increasingly part of large datasets containing web videos, the interactions considered are often relatively dissimilar and well-defined (e.g. a handshake and a hug). This puts the focus on dealing with the variations in the visual input, rather than subtle variations in the physical performance of the interactions. Also, this practice neglects issues with potentially ambiguous labeling such as in Figure~\ref{Figure1}(center). We deem an increased consideration of the coordination of body movements as a key requirement for successful application in more social settings, in which a multitude of subtly varying interactions may be encountered.

\begin{table*}[!t]
\caption{\label{tab1}Summary of datasets with footage type and quantity, number of action/interaction classes and actors}
\centering
\begin{tabular}{|l|c|c|c|c|c|c|}
\hline
Dataset & Footage type & Scripted & Sequences & Duration & Classes & Actors \\
\hline \hline
UT-Interaction & Outside recordings & Yes & 60 & 10-25s & 6 & 8 \\
\hline
TV Human Interaction & TV shows & Yes & 300 & 1-5s & 4 & 100+ \\
\hline
Hollywood2 & Films & Yes & 3,669 & 10-15s & 12 & 100+ \\
\hline
ShakeFive2 & Lab recordings & Yes & 153 with pose data & 5-10s & 5 & 33 \\
\hline
SBU Kinect & Lab recordings & Yes & 300 with pose data & 1-5s & 21 & 9 \\
\hline
AVA & Films & Yes & $\sim$57.6k & 15 min & 80 & 100+ \\
\hline
CMU Panoptic & Lab recordings & Partially & 65 multi-view with pose data & 10-15 min & N/A & 16 \\
\hline
SALSA & Inside recordings & No & 8 multi-view with sensor data & 30 min & N/A & 18 \\
\hline
Kinetics & YouTube videos & No & $\sim$500k & 10-15s & 700 & 100+ \\
\hline
Moments in Time & YouTube videos & No & $\sim$800k & 1-5s & 340 & 100+ \\
\hline
HACS & YouTube videos & No & $\sim$1.5M clips ($\sim$490k positive) & 2s & 201 & 100+\\
\hline
\end{tabular}
\end{table*}

\subsection{Survey overview}
% outline
The survey structure is as follows. Section~\ref{Section2} summarizes publicly available datasets. We then continue with an in-depth discussion of human-human interaction recognition literature. We distinguish between the more traditional methods based on hand-crafted features (Section~\ref{Section3}) and those based on deep learning (Section~\ref{Section4}). Finally, we discuss the limitations of the state-of-the-art and present promising avenues for further research.

\section{Datasets} \label{Section2}
\begin{figure*}[htb]
\centering
\includegraphics[width=\textwidth]{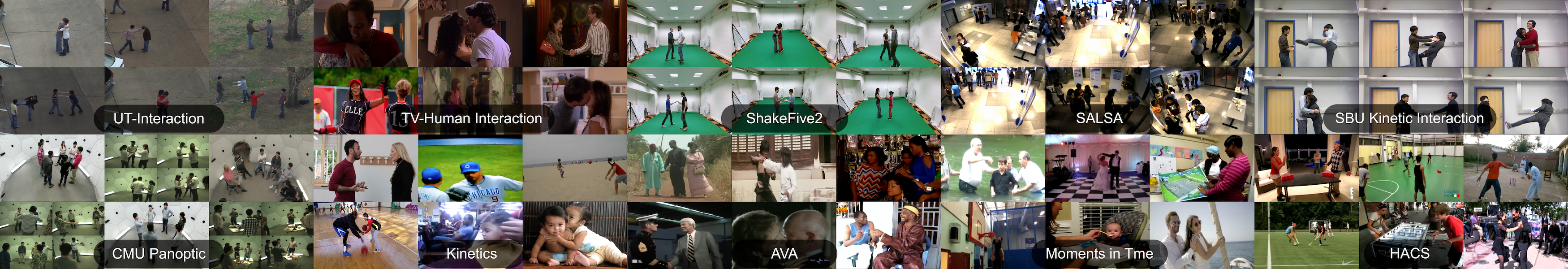}
\caption{Example video frames from different datasets depicting different interaction categories.}
\label{Figure2}
\end{figure*}

The availability of labeled datasets and the direct comparisons between methods generally lead to better understanding of the relative algorithmic advantages and limitations and, consequently, progression in performance. Compared to datasets available for individual action recognition (e.g., \citet{heilbron2015activitynet,kuehne2011hmdb, rodriguez2008action,soomro2012ucf101}), resources for human-human interactions are scarce. Most notably, the limited variation in viewpoint, application context and movement performance has hindered remarkable breakthroughs in the recognition of subtly different interactions such as those encountered in social settings. This section provides an overview of the most common datasets. Example frames appear in Figure~\ref{Figure2}. A summary of the datasets appears in Table~\ref{tab1}.

\subsection{UT-Interaction}
\textit{UT-Interaction} \citep{ryoo2010ut} contains 20 sequences and six interaction classes. With almost static background, limited occlusions and a fixed viewpoint, the classification difficulty is low. UT-Interaction is used as benchmark for many methodologies, ranging from bounding boxes techniques \citep{motiian2017online,shu2017concurrence} to bags-of-visual-words \citep{shariat2013new,nour2014human}. Some works have also addressed the detection of interactions in both space and time \citep{van2018hands}.

\subsection{TV Human Interaction}
The \textit{TV Human Interaction} dataset is composed of short video segments of four classes (\textit{handshake}, \textit{hug}, \textit{kiss} and \textit{high-five}), taken from popular TV series \citep{patron2010high,patron2012structured}. The dataset includes annotations of the upper bodies, head orientations and interaction labels for each person in the scene. Compared to \textit{UT-Interaction}, the video quality is higher, more different viewpoints and scenes are included and there is more variation in the number of people in the scene. All interactions are acted and the recording setting is highly controlled.

\subsection{Hollywood2}
\textit{Hollywood2} \citep{marszalek2009actions} also consists of clips from movies. Subtitles were used to align script data with the corresponding movie scenes. Despite the significant variation in the videos, the controlled nature of the movie domain limits generalization to more realistic domains. The four interaction classes are \textit{fight}, \textit{handshake}, \textit{hug} and \textit{kiss}.   

\subsection{ShakeFive2}
A collection of human interaction clips with complementary skeletal data was introduced by \cite{van2016spatio}. The videos are captured with fixed viewpoint and static background. The challenge of the dataset is in the similarity of the interaction classes (\textit{fist bump}, \textit{handshake}, \textit{pass object}, \textit{high-five} and \textit{hug}).

\subsection{SBU Kinect Interaction}
Additional depth data (RGB-D images), obtained from a Kinect sensor, is available in the \textit{SBU Kinect Interaction} dataset \citep{yun2012two}. It features eight two-person interactions: \textit{approach}, \textit{depart}, \textit{kick}, \textit{punch}, \textit{hand shake}, \textit{hug} and \textit{pass object}. The clips are segmented in time, with the interactions fully occupying the frame.

\subsection{CMU Panoptic}
\label{CMU}
The \textit{CMU Panoptic} dataset \citep{joo2015panoptic} is recorded in a large geometric dome with RGB and Kinect cameras distributed across the surface. The data are comprised of 480 synchronized video streams with additional pose information. Each clip depicts 3-8 people participating in social engagements: \textit{ultimatum}, \textit{prisoner's dilemma}, \textit{mafia}, \textit{haggling} and \textit{007-bang}. The activities are scripted but the interactions are genuine. No action classes have been defined but the participants closely interact.

\subsection{Kinetics}
\label{Kinetics}
The \textit{Kinetics} dataset \citep{carreira2019shortnote,kay2017kinetics} contains 700 video classes with approximately 600 videos per class. There are 11 interaction classes, including \textit{handshake}, \textit{hug} and \textit{massage feet}. The dataset is a collection of clips from YouTube videos. The video material is not professionally edited and features a large variety of background clutter, illumination settings and motion blur.

\subsection{Atomic Visual Actions (AVA)}
\label{AVA}
The AVA dataset \citep{gu2018ava} is composed of 15-minute segments from 432 movies. In addition to the labeling of clips for recognition, the interactions and actions of the actors within scenes are localized for tracking and detection tasks. The dataset contains 80 classes, including 13 interaction categories. The videos contain limited camera blur and most of the scenes have been shot with a still camera.

\subsection{Synergetic sociAL Scene Analysis (SALSA)}
\label{SALSA}
The SALSA dataset \citep{alameda-pineda2016salsa} contains 30 minutes of a poster presentation event, and 30 minutes of a cocktail party. In addition to camera views, the events have been recorded with various other sensors, including microphones and accelerometers. The data is richly annotated in terms of body and head orientation, and group membership. SALSA allows for the analysis of more social (group) interactions.

\subsection{Moments in Time}
\label{MomentsinTime}
The Moments in Time dataset \citet{monfort2018moments} is composed of three-second clips of events and activities. The dataset contains significant intra-class variation. Apart from common activity and interaction classes such as hugging and handshaking, some classes focus on group events such as dinning, baptizing or autographing.

\subsection{Human Action Clips and Segments (HACS)}
\label{HACS}
The HACS dataset \citep{zhao2019hacs} contains annotations of roughly 50k YouTube videos that correspond to 1.5M clips in total. The extracted two-second clips from the videos cover 201 classes, and also include negative samples that do not contain any action or interaction of interest, but are shot under the same image conditions. The dataset contains 23 interaction classes, mostly relating to sport activities.

\section{Recognition from handcrafted features}
\label{Section3}
Traditionally, the recognition of interactions from video starts with the representation of the scene and events as image features, and the subsequent classification of these features into an interaction class. Image features should be invariant to image conditions and interaction performance, while being sufficiently rich to deal with subtle differences between interaction classes.

% outline
We distinguish between local feature approaches that rely on salient points in the video, and template-based approaches that take into account regions in the video that roughly correspond to a person's body or body parts.

\subsection{Local features approach}
In general, local feature algorithms take a bottom-up approach by first detecting interesting points in a video, and then to aggregate detections over time and space to understand which behavior is being performed. These interesting points are selected locally, typically at edges or motion boundaries. Popular descriptors are based on Harris corners \citep{marin2013exploring,zhang2013recognition}, SIFT descriptors \citep{delaitre2010recognizing,lowe1999object} or optical flow \citep{yu2012propagative}. There is typically no direct correspondence between a point and a person or body part. As a consequence, factors such as camera motion, dynamic backgrounds and occlusions affect the presence of local features.

% 3D
%When additional depth information is available, e.g. from RGB-D recordings, local features can take into account depth gradients \citep{li2010action}. For the efficient mapping of 3D points from other viewpoints, \citet{xia2013spatio} considered the use of depth-sequences with the creation of a codebook.

% bow and fisher vectors
To increase the robustness of local descriptors, a distribution of points is usually described as a bag-of-words (BoW) or Fisher vector (FV) \citep{gao2016constrained,oneata2013action}. Instances of the same interaction class are assumed to have similar descriptors. To allow for a more complex distribution of the features, \citet{niebles2008unsupervised} construct a vocabulary using latent topics models. 

% spatio-temporal
Instead of modeling the trajectories of individual points, researchers have addressed the sequential nature of interactions by modeling the changes in the distribution of interest points over time. \citet{zhang2012spatio} use spatio-temporal phases to create a histogram of bag-of-phases. Each phase is composed of local words with specific ordering and spatial position. Instead of jointly mapping both dimensions, authors have addressed separation as well \citep{shariat2013new,tran2014activity}. The computed histograms represent similar features in single or multiple frames. Histograms of visual words have also been utilized by \citet{kong2012learning}. Here, the words derived from the quantization of the spatial-temporal descriptors were clustered to form a high-level representation of dyadic interactions, termed interactive phases. These phases include motion relationships such as the shaking of two hands. This idea has been extended to localize interactions by spatially clustering the phrases \citep{tran2013social}. To allow for variation in the temporal domain, \citet{prabhakar2012categorizing} model the causality of the occurrence of visual words.

% bow for patches / trajectories
Not all motions and attributes are informative, such as the positioning of the feet when performing certain greetings. \citet{kong2014discriminative} consider only body parts that characterize the interaction. Their method pools BoW responses in a coarse grid. This allows them to identify specific motion patterns relative to a person’s location. The level of detail of the analysis is limited by the granularity of the patches and the accuracy of the person detector. Additionally, they take into account the temporal nature of interactions by linking subsequent detections into trajectories. \citet{mohammadi2015violence} extend this approach by grouping the motion patterns as BoW vectors. Similarly, \citet{turchini2016understanding} introduce an approach to localize interactions from the trajectories of multiple local feature types. \citet{wang2013action} have introduced Improved Dense Trajectories (DT), a widely adopted way of finding and describing trajectories of points. In DT, a point is encoded as a combination of Histograms of Oriented Gradients (HOG), Histograms of Oriented Flow (HOF) and Motion Boundary Histograms (MBH). Points are linked over time.

% person and context
Local features can be used to isolate a person in video first. Extensive work has been done on the detection of humans from local features, encoded with HOG and HOF descriptors \citep{caba2016fast}. Once a person has been localized, the context of motions and actions of other people in the scene can provide useful cues for the recognition of their interactions. \citet{reddy2013recognizing} exploit the information obtained through a scene context descriptor which combines the location and surroundings extracted with optical flow and 3D-SIFT, based on the moving and stationary pixels. \citet{cho2017compositional} introduced the compositional interaction descriptor that takes into account the local, global and individual movement in video sequences. By linking local features to persons, we can describe their surroundings. \citet{lan2012discriminative} presented an Action Context (AC) descriptor that is based on connected action probability vectors of several people. Similarly, \citet{choi2014understanding} perform joint tracking, classification of the actions of an individual and the recognition of collective activities by considering bounding boxes of extracted local features.

\subsection{Template-based approaches}
% hog/hof - combi with local
%When a larger region in a frame is considered, we can address the recognition of a person's action from specific parts of the human body. Patches in an image can be described as HOG, HOF or MBH. HOGs describe the edge orientations within a grid of cells and encode spatial information such as a specific pose. HOFs are similar, but consider movement vectors from optical flow. Consequently, they encode motion direction within a spatial grid. Finally, MBH describes motion boundaries in a related way.

% general templates - hog and poselets
When applied to a single frame, a HOG descriptor can represent a characteristic pose. For example, a high-five interaction can be described as two people facing each other with outstretched hands that meet above their heads. This notion was adopted by \citet{bourdev2010detecting} to detect people engaged in specific actions, and was applied to human-human interactions by \citet{raptis2013poselet}. \citet{sefidgar2015discriminative} have formulated an implementation with discriminative key frames and their relative distance and timing within the interaction. Alternatively, \citet{sener2015two} formulate interaction detection as a multiple-instance learning problem to focus on relevant frames, because not all frames in an interaction are considered informative.

% hof
The motion around a characteristic pose can provide complementary information. \citet{van2014dyadic} combine HOG and HOF descriptors to encode the characteristic frame of a two-person interaction. \citet{yu2015fast} concatenate HOG and HOF descriptors and applied FV to make the detection linearly separable, thus allowing the model to concurrently utilize spatial and temporal features.
%While optical flow can be seen as the representation of the motion between two subsequent frames, tracklets describe the path of local key-points over longer time intervals. %\citet{mousavi2015analyzing} introduce Histogram of Oriented Tracklets (HOT) that summarize tracks of local key-points.

% combi with local
%Patches are sometimes combined with local descriptors to take advantage of potentially conjoint salient pose or motion information \citep{ji2016multiple}. \citet{yin2012small} employ 3D-SIFT to describe local motion events, but used a HOF to model the global motion in an image. Similarly, \citet{lathuiliere2017recognition} combined HOG descriptors and trajectory information from linked local features. Single-person and two-person interaction attributes such as ``two persons are standing side-by-side'' were calculated from these features.

% two-stage (face, body)
Instead of relying on interest points, we can first detect faces or bodies using a generic face or body detector \citep{patron2012structured,ryoo2011stochastic}. Given two close detections, interactions can subsequently be classified based on extracted features within the detection region \citep{ryoo2011stochastic}. Various attributes, including gross body movement and proximity, have been employed to classify the interaction. \citet{patron2012structured} also include the relative size and orientation of each person. \citet{khodabandeh2015discovering} consider clusters of similar frames based on proximity and appearance of pairs of people. They find that user feedback helps to increase the purity of the clusters, in turn improving the interaction classification. The drawback of this two-stage approach is that classification is sub-optimal when the person localization fails, for example when people partly occlude each other. This is a common situation, especially when people interact in close proximity.

% dpm
This issue is mitigated when employing Deformable Parts Models (DPMs) \citep{felzenszwalb2010object}. Here, an articulated object such as a person or multiple interacting people are modeled as a set of parts and deformations between them. This allows for more flexibility in the spatial layout of the parts. As such, parts that are generally well detected, e.g. a person's head, can be coupled with parts that are traditionally more challenging to detect, such as a lower arm. \citep{lu2015human} use a DPM as a prior to localize the rough outline of a person. Optical flow is then used to propagate the outline to subsequent frames. The resulting volume is then segmented into supervoxels to refine the person's outline in each frame, and classified as action. \citet{van2018hands} use interaction-specific DPMs with poselet parts \citep{bourdev2010detecting} to locate people in poses characteristic for a given interaction. Instead of encoding the orientation of (pairs of) limbs as poselets, DPMs can also include a larger number of articulations by using a mixture of parts \citep{yang2011articulated}. This approach has been used to describe the joint poses of two interacting people \citep{yang2012recognizing}. %\citet{hoai2014talking} extend the model to account for deformations in scale. Similarity with template examples of typical interactions are then used to rank the detection scores. The main constraint of this method is that it is used primarily for TV material, with an emphasis towards upper body movement.%

% temporal
%To account for more variation in the temporal performance of interactions, authors have introduced various methods. \citet{ji2017new} model the changes in HOG descriptors over time using a Hidden Markov Model (HMM). Based on the distance between two people, they consider the frame to be in the start, middle or end stage of the interaction. The HMM scores for the stages are fused for the final classification. The same rationale of splitting an interaction into phases has been adopted by \citet{cao2013recognize}, who address the task of classifying a sequence with potentially missing frames. 

% temporal dpms
While DPMs encode a particular pose or motion spatially only, extensions have been proposed to deal with the time-varying nature of human interactions. \citet{yao2014animated} focus on human-object interactions and capture the movement related to a key pose using a DPM and a linked set of motion templates that also correspond to different phases of the performance. \citet{tian2013spatiotemporal} have extended DPMs for action detection to model changes in pose over time. These formulations work well for the representation of coarse movements, but finer-scale movements are difficult to model because the motion is not linked to specific parts of the body.% An extension of DPMs to include part deformations not only in space and scale, but also in time is presented in \citet{van2018hands}. This enables the detection, in both space and time, of interactions that are characterized by a single key pose or key movement phase. \citet{tran2012max} also address a localization task but consider linking regions over time based on HOG and HOF in a structured learning approach. A max-path algorithm is used to find the optimal volume that contains the action in space and time.%

\section{Interaction detection from learned features}
\label{Section4}
The hand-coded feature descriptors described in Section~\ref{Section3} focus on local or global spatial or spatio-temporal information. The manual selection of descriptors leaves room for improvement because the process is agnostic to the specific classification task, application domain or class of behaviors.

% cnns
Based on the introduction of multiple convolutions by \citet{lecun1998gradient}, Convolutional Neural Networks (CNNs or ConvNets) have been used for classification tasks of both image and video data. CNNs allow for the simultaneous training of a classifier, and the automated selection of informative features. Consequently, they can overcome the issue of sub-optimal feature selection. While multiple convolution kernels allow for the selection of a wide range of image or video features, the stacking of consecutive convolution operations allows for a hierarchical extraction of complex features \cite{simonyan2014very}. Typically, the characteristics extracted in the first layers of the network correspond to low-level features such as edges and simple textures. Deeper layers of the network are targeted towards the extraction of higher-level features.

% requirements
Methods based on neural networks have shown notable improvements in human action and interaction classification tasks. Deep learning benefits from extensive amounts of data without saturation in the accuracy rates equivalent to the data growth rate. This allows deep learning architectures to generalize their feature assumptions, based on the utilization of all potential information in images and videos, rather than being limited to a predefined set of features, as in the hand-crafted methods. 

% aim of section
The purpose of this section is to present neural network architectures for human interactions that operate on single frames. We then show how temporal information can be incorporated in the convolutions and finally discuss recurrent models.

%\begin{figure*}
%\centering
%\includegraphics[width=\textwidth]{Grad-cam-kiss.png}
%\caption{Highlighted regions demonstrate the focus of the network on the most informative parts of each fifth frame using the Grad-cam visualization \citep{selvaraju2016grad}. We have used Transfer Learning based on the Inception V3 \citep{szegedy2015going} architecture with the network weights pre-initialized on ImageNet. We retrained the last two fully connected layers and the prediction layer on the TV Human Interaction dataset \citep{patron2010high}. The code is available at: \url{https://github.com/alexandrosstergiou/Inception_v3_TV_Human_Interactions}}
%\label{Figure3}
%\end{figure*}

\subsection{Single frame networks}
CNNs have been used to classify actions and interactions in single frames \citep{asadi2017survey,bilen2016dynamic,gkioxari2015contextual}. Similar to the use of handcrafted features, the focus is on characteristic joint poses. To extend this methodology to sequences of images, several approaches have been proposed.

% Early, Late and Slow Fusion
Based on the classification of individual frames, \citet{karpathy2014large} proposed three techniques to fuse the scores of multiple frames using different convolutional configurations. In the Early Fusion strategy, the input of the network is a stack of subsequent frames. Late Fusion combines the convolutional features of the first and last frames of a sequence in the final, fully connected layers. Slow Fusion is a combination of these two approaches, that empowers a progressive fusion over frames and activation maps, with the extension of convolutional layer connections through time. All three approaches are limited in their capability to deal with subtle temporal variations between classes, and large intra-class variations. It is a challenge to deal with these variations as they have to be modeled from the typically modest number of training videos.

% transfer learning
To partly mitigate this issue, authors have investigated the use of Transfer Learning \citep{bengio2011deep,bengio2012deep,caruana1998multitask,pan2010survey,yosinski2014transferable}. This is a process in which the network is first trained on a large dataset with general examples, and subsequently re-purposed for another, more specific, classification task. In general, this means that the deeper layers are retrained for the specific domain. Consequently, fewer parameters need to be learned for the novel domain, which reduces the risk of overfitting.

\subsection{Motion-based and stream networks} \label{Section4.2}
% two-Stream CNNS & multiple streams
% Two-stream with pose stream zolfaghari2017chained
Two-stream CNNs combine regular images and optical flow images as input \citep{simonyan2014two}, and are an alternative approach to model temporal information. The rationale is that still images encode the pose of an interaction, while the optical flow provides information about the motion. The network consists of two streams, branches in the network structure. The spatial-based CNN is trained on individual video frames, and the temporal stream CNN takes as input stacked optical flow fields from multiple frames. The results from the two networks are concatenated with late fusion. Different information fusion methods for each stream were explored by \citep{park2016combining}. \citet{wang2016temporal} added a Temporal Segment Network (TSN) to the two-stream CNN architecture, applied on sporadically sampled fragments from the video, thus making a prediction on each of the snippets independently. The predicted class is then the `point of agreement' between the video segments. This method capitalizes on information from small temporal segments rather than using the video as a single input. Following the use of selected frames \citet{diba2017deep} also proposes a representation and encoding of the sequence features in a Temporal Linear Encoding (TLE) layer, after the convolution feature extraction is performed. It is based on the aggregation of appearance features from each of the individual temporal fragments. Works have also included the use of depth data as stream inputs \citep{garcia2018modality} in which features from the depth stream are distilled in order for the depth stream to be simulated at test time as the test data does not include this supplementary modality.

% links between the streams
Inputs in the two-stream CNN are processed independently and only fused as a last step. This approach prevents the exchange of information between the streams. As such, it is not possible to develop attention mechanisms that focus on specific parts on the input in either stream. One way of establishing these links is by using skip connections of Residual Networks \citep{he2016deep,hara2018can} and additional shortcut connections between convolutional layers of the motion stream to the spatial stream. This provides benefits in optimizing the network architecture and increasing the network depth \citet{feichtenhofer2016spatiotemporal}. Residual learning enables the model to avoid degradation in deep structures, which relates to the saturation of accuracy followed by a significant drop when optimizing the parameters as layers of the network are not able to effectively learn the identity map and instead ``threshold'' to zero mappings.

% specific architectures
Recent advances in reinforcement learning and evolutionary algorithms have contributed to a reduction in human supervision for creating robust network architectures \citep{zoph2016neural}. This trend has further enabled the construction of architectures for specific tasks rather than general architectures \citep{Zoph_2018_CVPR}. With an increasing number of options for layers and connections, such techniques are welcome to avoid the slow research progress due to extensive parameter testing.

% r-CNN
Typically, a human interaction does not occupy the entire frame. So instead of taking the entire image or image sequence as an input, the region corresponding to the actual interaction can be identified first and used as input. One technique that takes this two-step approach is Regional CNNs (R-CNN) \citep{girshick2014rich}, that classify each region with a category-specific linear SVM. Notably, \citet{peng2016multi} demonstrated a multi-regional two-stream R-CNN which uses a region-of-interest fusion layer for both appearance and motion models. Region-focused, stream-based models have also been used by \cite{tran2017two}, who introduce cross-connections from the temporal to the spatial stream. These include convolutions that reduce the dimensionality of the temporal activation maps. The hierarchical model for features has also been used for the creation of action tubes \citep{gkioxari2015finding}: spatio-temporal volumes centered on the performance of a particular action or interaction. Here, region proposals are found based on motion-appearance cues extracted with a two-stream CNN. The notion of using tubes for the representation of motion has also been adopted for different body parts by \citet{mavroudi2017deep}. \citet{saha2016deep,hou2017tube} have also implemented a model based on action tubes and R-CNNs as well as connections between the spatial and temporal models.

% R-CNN improvements
Adaptations to regional CNN models have been created by \citet{gkioxari2015contextual} and \citet{mettes2017spatial} to include multiple regions per example. The primary region contains the main actor or actors, while secondary regions are based on contextual cues of the scene. Similarly, \citet{wang2016two} used a two-stream semantic region-based CNN (SR-CNNs) as an extension of Faster R-CNNs \citep{ren2015faster}. The idea of using multiple independent or dependent regions for various cues, and using separate streams to encode the input, also allows to focus on discriminative regions such as a hand of one person that touches the body of another \citep{singh2016multi,miao2017multimodal,tu2018multi,wu2016multi}. Typically, such regions complement each other.

%3D-CNNs
Instead of treating the image and motion aspects of a video in separate streams, a video sequence can be represented as a 3D volume that is composed of stacked frames. \citet{baccouche2011sequential} and \citet{ji20133d} use 3D convolutions to simultaneously encode the spatial and temporal features of such a volume. This approach is essentially an extension of the standard 2D convolutions to 3D. The resulting feature maps encode informative spatio-temporal patterns in the video volume. \citet{tran2015learning} presented the C3D architecture and demonstrated its superiority over 2D CNNs. 3D convolutions can also be used concurrently with a two-stream network. \citet{carreira2017quo} have introduced a fusion of these two methodologies, two-stream inflated 3D-CNNs (I3D), that adds a temporal dimension to the kernels of both convolutional and pooling layers. The work considers the creation of two I3D models that are applied to static image and optical flow inputs, and thus allows the 3D-CNNs to benefit from the additional information of motion patterns in optical flow streams. Spatio-temporal networks can be used as a base architectures to extend the type of information processed such as queries for people regions \citep{Girdhar_2019_CVPR}, position and motion \citep{choutas2018potion} and feature neighborhood correspondence across time \citep{cao2019gcnet,wang2018non}.

% Pseudo, 2+1D and other spatio-temporal block variants
The larger number of parameters in 3D convolution blocks and, consequently, the demand of larger datasets for 3D-CNNs to train, have motivated the introduction of alternative convolution blocks. Notably, \citet{qiu2017learning} have proposed three supplementary blocks with different configurations of a single 2D convolutional kernel for the extraction of appearance information per frame and a temporal kernel responsible for the changes of pixel values over time loosely inspired by the separable convolutions of 2D-CNNs \citep{chollet2017xception,howard2017mobilenets}. This idea has also been used to separate spatio-temporal kernels into purely spatial and purely temporal ones by \citet{tran2018closer} with the introduction of (2+1)D convolution blocks. Others have fused both solely-spatial and spatio-temporal convolutions in an effort to emphasize the spatial signal \citep{zhou2018mict}. \citet{chen2018multifiber} have also proposed the slicing of convolutional blocks in sets of fibers that are processed in parallel by the model. This significantly reduces the computation overhead, owing to the decreased size of the activation maps produced by each operation at each fiber.

%All these works follow the notion that the further decomposition of the model, in the context of a challenging problem such as human interactions or actions, in sub-problems, can aid the better analysis of the task and the optimization of it.

\subsection{Recurrent networks}
While CNNs can recognize image components and learn to combine them to classify different classes, they lack the ability to recognize patterns across time. Stream-based networks and 3D convolutions can take into account motion, but do not explicitly deal with variations in the temporal performance of an action or interaction. An alternative approach is to use recurrent neural networks (RNNs) that model temporal patterns. The key idea is to use some form of recurrence in the network that allows the persistence of information through sequences of inputs. Thus the temporal variations in videos can be efficiently modeled alongside to the spatial variations.

% use of rnns
Recurrent neural networks have been effectively used as a supplementary architecture to CNNs for extracting temporal features. In such architectures, spatial information is extracted though CNNs and is then passed to recurrent networks for learning the temporal characteristics of each interaction class \citep{bagautdinov2017social, deng2016structure}. \citet{zhao2017two} proposed an approach based on the normalization of each layer of the network with batch normalization \citep{ioffe2015batch}. The architecture is combined with a 3D-CNN using a two-stream fusion of the RNN and CNN. The use of multiple recurrent networks has also been extended to include tree structures (RNN-T) \citep{li2017adaptive}, to perform a hierarchical recognition process in which each RNN is responsible for learning an action instance based on an Action Category Hierarchy (ACH). This allows for the distinction between very dissimilar classes high in the hierarchy, while subtle differences between related classes such as a handshake and a fist bump are dealt with in the lower nodes.

% lstms
Recurrent Neural Networks suffer from vanishing gradients. This issue causes the updates in the network weights of the top layers to gradually diminish as the number of data-processing iterations increases. This hinders learning the temporal parameters effectively. To overcome this issue, Long Short-Term Memory (LSTM) RNNs \citep{hochreiter1997long} have been introduced that include additional `memory cell' modules that decide whether to keep the processed information. As such, they are capable of maintaining information over longer periods, which allows them to learn long-term dependencies \citep{chung2014empirical}. This is essential for the modeling of interaction classes as the distinctive information is often present in different phases of the interaction.

% lstm temporal convolutions
\citet{donahue2015long,li2018videolstm, varol17} have shown that the combination of convolutions and long-term recursions performs well for recognition tasks in videos. \citet{donahue2015long} was effective in both image and video description by directly connecting powerful feature extractors such as CNNs with recurrent models. Similarly, \citet{baccouche2011sequential} extracted features from the 3D-CNN architecture and extended the work to a two-step recognition process with a LSTM. The first step was the use of 3D convolutions for the extraction of spatio-temporal features. The second step is based on these learned features that are passed to the LSTM so the model can make predictions on the entire video sequence. As such, the network can benefit from both short-term and long-term temporal information.

% highway networks
Besides LSTMs, Highway Networks are an alternative solution to the vanishing gradient problem \citep{srivastava2015training}. These networks allow for the direct passing of information through so-called highway modules that connect layers of the architecture similarly to LSTM's adaptive gating mechanism. \citet{zilly2017recurrent} have extended this approach to include the spatial dimensionality in the information highways inside recurrent transitions.

% attention/selection in LSTMs
Because the discriminative information of an interaction is typically found in selective parts of the input, several approaches have addressed methods for selection. In line with the multi-stream approaches (Section~\ref{Section4.2}), \citet{wang2017recurrent} have implemented LSTMs that consist of three branches that deal with person action, group action and scene recognition. This work is inspired by \citet{gkioxari2017detecting}, who focused on human-object interactions instead. Multiple recurrent modules can be used to analyze human interactions. For example, \citet{yan2017predicting} built a model from three attention-specific LSTMs that use information from each of the two interacting actors and the overall scene of each example. Similarly, \citet{Si_2019_CVPR} also included spatio-temporal focused LSTMs, through a temporal hierarchy, for increasing the temporal receptive field of the network and allowing the exploration of co-occurring features in space and time. \citet{ibrahim2016hierarchical} presented a two-stage temporal model in which LSTMs are used to analyze each person in the scene while their combined outputs synthesize the relationship between them. \citet{srivastava2015unsupervised} created an Encoder-Decoder architecture, in which the encoder LSTM maps input sequences to a delineation of specified length. The decoder LSTM then either reconstructs the inputs or creates predictions for future examples. The motivation of the work is to capture all information required to reproduce the input and therefore to select the most important features. This is achieved by minimizing the loss of the constructed sequence from the decoder LSTM and the actual input sequence. For example, in an interaction video, the decoder would focus on modeling the movement of the hands if the interaction is a handshake, or focus on the upper bodies if the interaction is a hug.

% Skeletal approaches
Of increasing importance for interaction recognition is the use of skeletal data, or poses. Pose data is a compact representation that is invariant to many typical image factors such as partial occlusions, low resolution and viewpoint. Consequently, the focus is mainly on modeling the temporal dynamics. Often, pose information can be regarded as a complementary input. For example, \citet{gammulle2017two} have created a spatio-temporal two-stream architecture with an addition of a LSTM with both frames and optical flow working as an attention mechanism. Attention mechanisms have also been used with pose information in recurrent structures to learn pose-related features in each time step \citep{du2017rpan}. This permits the analysis of the action from the collection of the per-frame human poses. Moreover, based on alternatives to LSTMs, \citet{liu2016spatio} have introduced gating mechanisms for creating a spatio-temporal LSTM (ST-LSTM). Given skeletal data in a tree-like structure, each ST-LSTM unit corresponds to a joint and receives spatio-temporal information from the previous and its own node. The new gating mechanism predicts the possible input based on the generated probabilities and compares it to the actual input. They implement the idea of assimilating the sequential input of videos by adjusting the effects on the context-based information stored in the network by allowing to analyze the data at each step and to decide when to update, remember of forget the contents in the memory cell with a tree-like representation of the skeleton.

Skeletal data have also been used by \citet{zhu2016co} in a fully connected LSTM model including internal gates, outputs and neurons that could be dropped by the network. \citet{si2018skeleton} have proposed a combination of networks. The first network analyzes spatial information between frames by capturing the relationships between skeletal joins, while the second network focuses on the dynamics and the detailed temporal features that define each example. Other extensions include Lattice-LSTM (L$^{2}$STM) that enhances the capability of the memory cell to understand motion dynamics of the video sequence through individual local patterns, by leveraging both image and flow information extracted from a CNN classifier \citep{Sun_2017_ICCV}. Since there might be different patterns for different body parts and phases in the interaction, LSTMs have been adapted to consist of part-based sub-cells to model the long-term motion of key body parts \citep{du2015hierarchical, shahroudy2016ntu}. Because these models break down the interaction into meaningful blocks of motion, they can be used as the basis to learn a repertoire for action and interaction as shown by \citet{Shi_2019_CVPR}. They introduce a directed acyclic graph (DAG) representation for the information of joints, bones and their relationship. Other approaches have targeted the dependencies among joints \citep{Li_2019_CVPR} where information is also separated to actional links based on movement and structural links based on joint locations. By partitioning the interaction into sub-parts, these approaches can further reduce training cost and lead to the distinction between subtly different interaction classes.

\section{Discussion} \label{Discussion}

\begin{figure*}
\centering
\includegraphics[width=\textwidth]{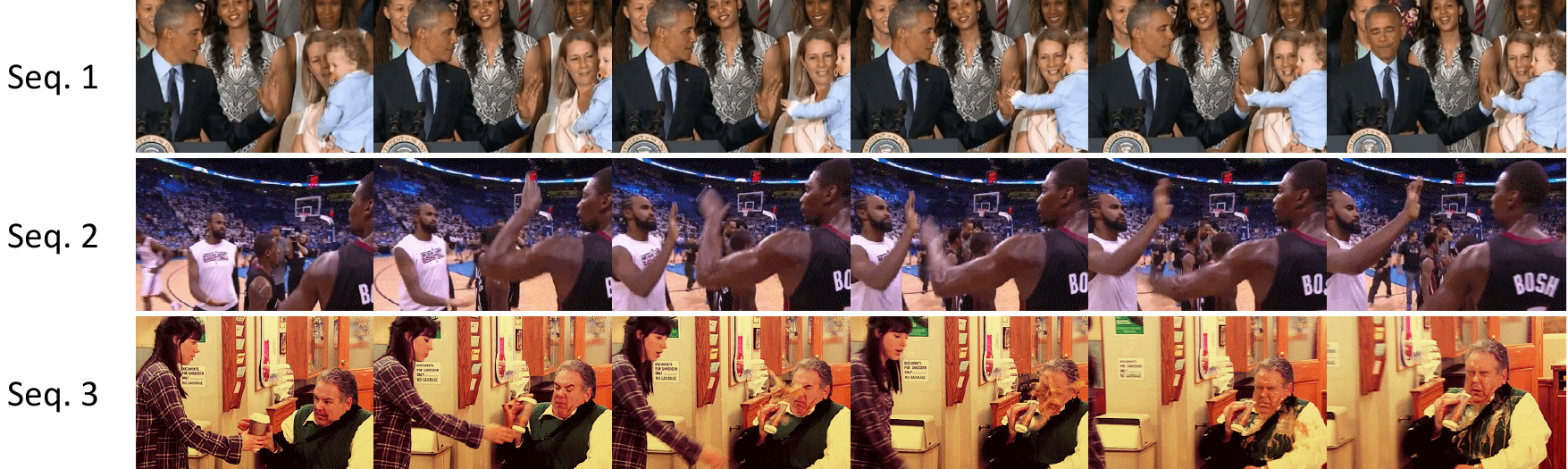}
\caption{Examples of ambiguous interactions. Sequence 1 shows that ambiguity can arise from an unexpected outcome: a high five that ends in holding hands. In Sequence 2, there is no contact between the two persons but their motivation for a high-five is apparent. There is comical intent in the interaction in Sequence 3. The comprehension of this scene requires deeper understanding of the interactions.}
\label{Figure5}
\end{figure*}

% summary and advantages of cnns
The past decades have seen impressive progress in the automated understanding of human behavior in videos. With the introduction of learned feature approaches such as CNNs, we can now analyze videos recorded in unconstrained settings. Consequently, there is a focus on more realistic video material. The result of initial works on specially recorded benchmarks datasets have largely saturated. In the meantime, we have begun to address sustained, natural human interactions in a social context. This opens up a host of applications, from more intelligent video indexing to smart surveillance.

% challenges
In Section~\ref{Section1.2}, we discussed a number of challenges. The introduction of learned feature representations has alleviated some of the issues when dealing with variations in recording setting, person appearance and, to a lesser extent, viewpoint. The decoupling of the visual and temporal aspects of human interactions, for example using LSTMs \citep{alahi2016social}, has allowed researchers to focus more on the dynamics of interactions. Still, the promise of understanding social interactions directly from video is far from being met. Below, we discuss limitations of the state-of-the-art and highlight current trends and future directions.

% computation requirements
\paragraph{Training scenarios with less data}
Advances owing to CNNs come at a cost because learned feature representations require large amounts of relevant training data. While the datasets that focus on human interactions are still increasing in the number of classes and available videos, it will remain hard to harvest such datasets. Some works have exploited synthetic data generators to increase the amount and variation of the training data \citep{Chen_2017_ICCV,shotton2013real}. The generation of the data can also explicitly be part of the training process. Generative Adversarial Networks (GANs, \citet{goodfellow2014gan}) contain a generative and a discriminative model that are jointly optimized. Recent work on the walking motion of pedestrians demonstrates the efficacy of the technique to model social behavior \citep{gupta2018socialgan}. It remains to be investigated to what extend these results generalize to less-constrained interactions. Another line of approach is to use transfer learning \citep{weiss2016survey}, to learn the parts of the network that deal with the lower-level aspects of the input from more general and more widely available training data. Despite these partial solutions, there typically is relatively few relevant data available given the complexity of the classification problem.

%This introduces technical issues. Most notably, as shown by \citet{goodfellow2014explaining, szegedy2013intriguing, su2017one} convolutions are susceptible to adversarial noise, as they can misclassify examples that look similar to humans but have slightly different pixel values.%

% limited repertoire, no subtle differences
\paragraph{Increasing interaction class repertoire}
Current work on the analysis of human interactions is limited by a relatively coarse division into behavior classes such as a handshake or a hug. Often, there is much more information contained in these interactions and humans have little difficulty identifying an awkward hug from a heartfelt one. \textit{Semantically}, such interactions are very different. Yet, they can be visually very similar. With an increased focus on realistic human interactions comes a need to be able to distinguish between a larger number of classes, each of which might only subtly differ from others. These differences might originate from temporal aspects such as the coordination in time, but also from differences in poses or orientation. Completely separating the visual aspect from the temporal characteristics is likely to be sub-optimal. We consider the use of recurrent networks with more sophisticated gating functions as a promising trend.

The current practice is to consider an interaction as belonging to a single class only. But human behavior is often more open to subjectivity, and a less strict separation into classes could be beneficial for the generalization. The work on overlapping labels or behavior hierarchies (e.g., \citep{frosst2017distilling,yeung2018everymoment}) is promising because it facilitates the focus on distinctive patterns at different levels of granularity, dependent on the type of interaction. A shift away from the one-vs-all classification can additionally facilitate the introduction of loss functions that take into account how related, visually or semantically, interactions are.

% units of motion, pose and coordination
\paragraph{Units of interaction}
Predominantly, interactions are classified directly based on the input. Some works have considered semantic mid-level features such as the action of an individual (e.g., \citet{lan2012discriminative, sefidgar2015discriminative}) or the action of a body part (e.g., \citet{cheron2015p,kong2012learning,tian2015deep}). Such methodologies bring some invariance in the representation, and can be learned per person. This effectively removes some of the dependencies and can facilitate the modeling of interactions as spatio-temporal patterns of these mid-level features. This approach can even be extended to deal with interactions for which no, or very little, training data is available. Specifically for human-human interactions, the coordination of pose and motion is crucial to distinguish between subtly different classes \citep{van2018hands}. Mid-level representations should take into account this coordination in both space and time, such as the distance and orientation between people, or the relative placement of a hand on the other's shoulder. Recent work on capsules by \citet{hinton2018matrix,sabour2017dynamic} appears promising in this respect. These works have shown great potential for accurately learning the pose of an object and constructing a hierarchy of parts enabling the understanding of features that is specific to a class. As such, geometric relations can be modeled in detail. An additional advantage is that capsules can be parallelized \citep{goyal2017accurate}, which limits the computational requirements.

% skeleton data
\paragraph{Role of skeleton data}
Human poses are one particular form of mid-level representation. We foresee an increased role of skeleton data, both during training and as additional input modality. Temporal patterns of interactions can be learned from skeleton data directly without having to take into account factors such as viewpoint and person appearance. Especially when units of interactions can be defined, pose and motion for an individual, as well as the coordination between people can be readily analyzed from skeleton data. Recent advances in human pose estimation from images and video (e.g., \citet{carreira2016human,insafutdinov2017arttrack,Yang_2017_ICCV}) have paved the way for effective pose-based attention mechanisms. While the computational requirements of the pose estimation task are significant, the benefit for the recognition of interactions has also been demonstrated \citep{du2017rpan,liu2016spatio}.

% detection vs. classification
\paragraph{Detection and classification}
The research on the automated analysis of human interactions has predominantly focused on recognition rather than detection. This means that interaction labels are usually not assigned to a region but to the image or video sequence as a whole. Rather, the understanding of human behavior would benefit from a link between person and interaction class. This permits us to say who interacted with whom, when. Especially in sustained or repeated social encounters, for example in public spaces, knowing the actors that interact would increase the efficacy of the analysis. A few works have addressed interaction detection (e.g., \citet{van2018hands,tian2013spatiotemporal}) but usually in a two-step approach by first detection humans (e.g., \citep{patron2012structured}) and then considering their interactions. Especially in more crowded settings where partial occlusions are more common, such an approach is more likely to fail. An approach that focuses on the distinctive parts of the interaction is therefore favorable.

% intent vs. observation, stories, social context
\paragraph{From observation to understanding}
Finally, we see much potential in leveraging the recognition of interactions to the understanding of interactive human behavior. While the analysis of the observations is an essential step to understanding video contents, it often is not sufficient for our common use and demands. Often we are looking for anomalies, deviations from common practice. For example, Sequences 1 and 2 in Figure~\ref{Figure5} show interactions that are difficult to recognize but are more likely to be of interest to a user. Descriptive units of interactions can be instrumental in modelling anomalies. Commonly, it is the context of the behavior that is more descriptive, or gives a different meaning to our interactions. When a person is observed pushing another, it could be a playful instance between two friends or an actual act of violence. Longer-term analysis of the actors, their roles or relation to each other and knowledge of social and cultural norms can help in providing a deeper understanding of the observed social behavior. In particular, the understanding of the intentions of a person can help to analyze what a person is doing, instead of focusing on how that is achieved.

When looking at videos, we should deviate from the current agnostic perspective and treat videos not as sequences of images but as visual representations of social behavior. We foresee that datasets that target a more constrained setting, yet contain a wealth of social behavior (e.g., \citet{alameda-pineda2016salsa}) are used as a step-up towards more generalized understanding of interactions from video. We identify a particular need for such datasets.

% closing
We are just scratching the surface when it comes to really understanding social behavior from video. But with the solid state-of-the-art performance in the analysis of interactions from videos, and the promising directions of research to deal with the current limitations, we expect that great strides can be made to close to the gap to the automated understanding of human interactions.

\section{Acknowledgments}
This publication is supported by the Netherlands Organization for Scientific Research (NWO) with a TOP-C2 grant for ``Automatic recognition of bodily interactions'' (ARBITER).

\bibliographystyle{model2-names}
\bibliography{main}

\end{document}